\documentclass{article}
\usepackage{ijcai26}

\usepackage{times}
\usepackage{soul}
\usepackage{url}
\usepackage[hidelinks]{hyperref}
\usepackage[utf8]{inputenc}
\usepackage[small]{caption}
\usepackage{graphicx}
\usepackage{amsmath}
\usepackage{amsthm}
\usepackage{booktabs}
\usepackage{algorithm}
\usepackage{algorithmic}
\usepackage[switch]{lineno}
\usepackage{comment}
\usepackage[table]{xcolor}
\definecolor{lightblue}{RGB}{220,235,250} 
\definecolor{greenblue}{RGB}{0,150,136}

\title{\textit{ArogyaSutra}: A Multi-Agent Framework for  Multimodal Medical Reasoning  in Indic Languages}

\author{
Tanmoy Kanti Halder$^{1,3}$\footnotemark[1]
\and
Akash Ghosh$^1$\footnotemark[1]
\and
Subhadip Baidya$^2$
\and
Arijit Roy$^1$
\and
Sriparna Saha$^1$
\affiliations{
$^1$Indian Institute of Technology Patna\\
$^2$Indian Institute of Technology Kanpur\\
$^3$Prasannadeb Women's College
}
\emails{
\{tanmoy.zx, subhadipbaidya23, akashghosh.ag90\}@gmail.com,
\{sriparna, arijitroy\}@iitp.ac.in
}
}

\begin{document}

\maketitle

\begin{abstract}

Multimodal Large Language Models (MLLMs) have shown promising reasoning capabilities in general domains, yet their performance remains limited in specialized settings such as healthcare, especially in multilingual and low-resource scenarios. This gap is critical in regions like rural India, where patients often express complex medical queries in native Indic languages and rely on multimodal inputs such as medical images. Existing English-centric MLLMs struggle to support such use cases, limiting equitable access to AI-driven healthcare assistance. To address this challenge, we introduce \textbf{ArogyaBodha}, a large-scale multilingual multimodal medical question-answer dataset constructed from eight heterogeneous sources, covering 31 body systems, six imaging modalities, and 21 clinical domains across English and seven major Indian languages. We further propose \textbf{\textit{ArogyaSutra}}, an actor-critic-based multi-agent framework that integrates tool grounding with dual-memory mechanisms for step-wise, reasoning-aware decision making, and uses stored actor-critic simulation trajectories for distillation. Experiments show that our dataset and framework improve multilingual medical reasoning accuracy across all Indic languages, with ablations validating the contribution of each component. The source code and dataset are available at: \textit{\url{https://iitp-cse.github.io/ArogyaSutra/}}.

\end{abstract}

\section{Introduction}
Multimodal large language models (MLLMs) have demonstrated strong performance on vision--language tasks like multimodal summarization, retrieval, visual question answering, etc and have been adapted for medical applications via supervised fine-tuning on curated multimodal instruction datasets~\cite{li2023llava,jiang2024med,acharya2025m3retrieve}. However, most medical MLLMs rely on a direct-prediction paradigm, producing short answers that fail to capture the interleaved image--text reasoning required in real clinical settings. While reinforcement learning approaches such as GRPO aim to elicit reasoning behaviors, their improvements remain constrained by the underlying base models and do not introduce fundamentally new reasoning paradigms~\cite{guo2025deepseek}. Although chain-of-thought supervision could enhance medical reasoning, curating high-quality medical CoT data is costly and lacks standardized evaluation, motivating the need for methods that can both generate and rigorously evaluate step-by-step multimodal reasoning for complex decision making.

\par

Multimodal Medical reasoning research remains largely English- and Chinese-centric, leaving mid- to low-resource Indic languages severely underrepresented and resulting in uneven reliability across linguistic communities~\cite{qin2025survey}. While cross-lingual transfer provides partial relief, open-ended multimodal medical reasoning in Indic languages continues to suffer from two persistent failure modes: \emph{degraded logical fidelity} and \emph{unstable language behavior} ~\cite{onyame2026cure}. Existing efforts to improve reasoning through domain-specific supervision are further constrained by benchmarks that are largely monolingual, text-centric, and closed-form, providing limited insight into multilingual multimodal reasoning and cultural consistency~\cite{costa2022no,surana2026viraasat,maji2025drishtikon}. As multimodal LLMs are increasingly adopted for clinical education and decision support in diverse linguistic settings, systematic evaluation of step-by-step multimodal reasoning and language fidelity in Indic languages is essential for fairness, reliability, and real-world applicability.\par

To address this gap, we introduce \textbf{ArogyaBodha}, a large-scale multimodal medical reasoning dataset and benchmark curated from eight diverse medical sources. \textbf{ArogyaBodha} spans seven major Indian languages alongside English and covers multiple imaging modalities, encompassing 31 body systems and 21 clinical domains. Each instance consists of an expert-verified multimodal medical query paired with a single unambiguous ground-truth answer, enabling reliable evaluation of both logical reasoning accuracy and language consistency, as well as systematic analysis of cross-lingual generalization under clinically grounded constraints as shown in  Table \ref{tab:dataset_summary}.

To enable reliable multimodal medical reasoning in Indic languages, we propose \textbf{\textit{ArogyaSutra}}, an actor--critic-based multi-agent framework that combines tool-based visual grounding with dual-memory mechanisms for step-wise, reasoning-aware decision making over image--text inputs. At each reasoning step $t$, the \textbf{Actor}, implemented as a multimodal language model, processes medical visual inputs and an Indic-language query, invoking lightweight grounding tools ~\cite{Zhou2025-nx} (e.g., zoom/crop, edge detection, depth analysis, and region-level detection) to extract clinically relevant evidence. Rather than directly producing a final answer, the Actor predicts intermediate semantic reasoning steps. The framework is supported by a dual-memory design: \emph{long-term memory} summarizes prior reasoning steps, linguistic formulations, and identified errors (from steps $1$ to $t\!-\!1$), while \emph{short-term memory} captures the most recent prediction error and feedback. The \textbf{Critic} evaluates the Actor’s outputs for both medical correctness and language consistency, providing targeted corrective feedback—delivered in English for linguistic errors and in the corresponding Indic language otherwise—which is used to update the memory modules. Through this iterative actor--critic interaction, {\bf \textit{ArogyaSutra}} enables transparent, medically grounded, and linguistically stable reasoning, supporting more trustworthy diagnostic explanations across diverse Indic languages.

\begin{figure*}[t]
    \centering
    \includegraphics[width=0.75\textwidth]{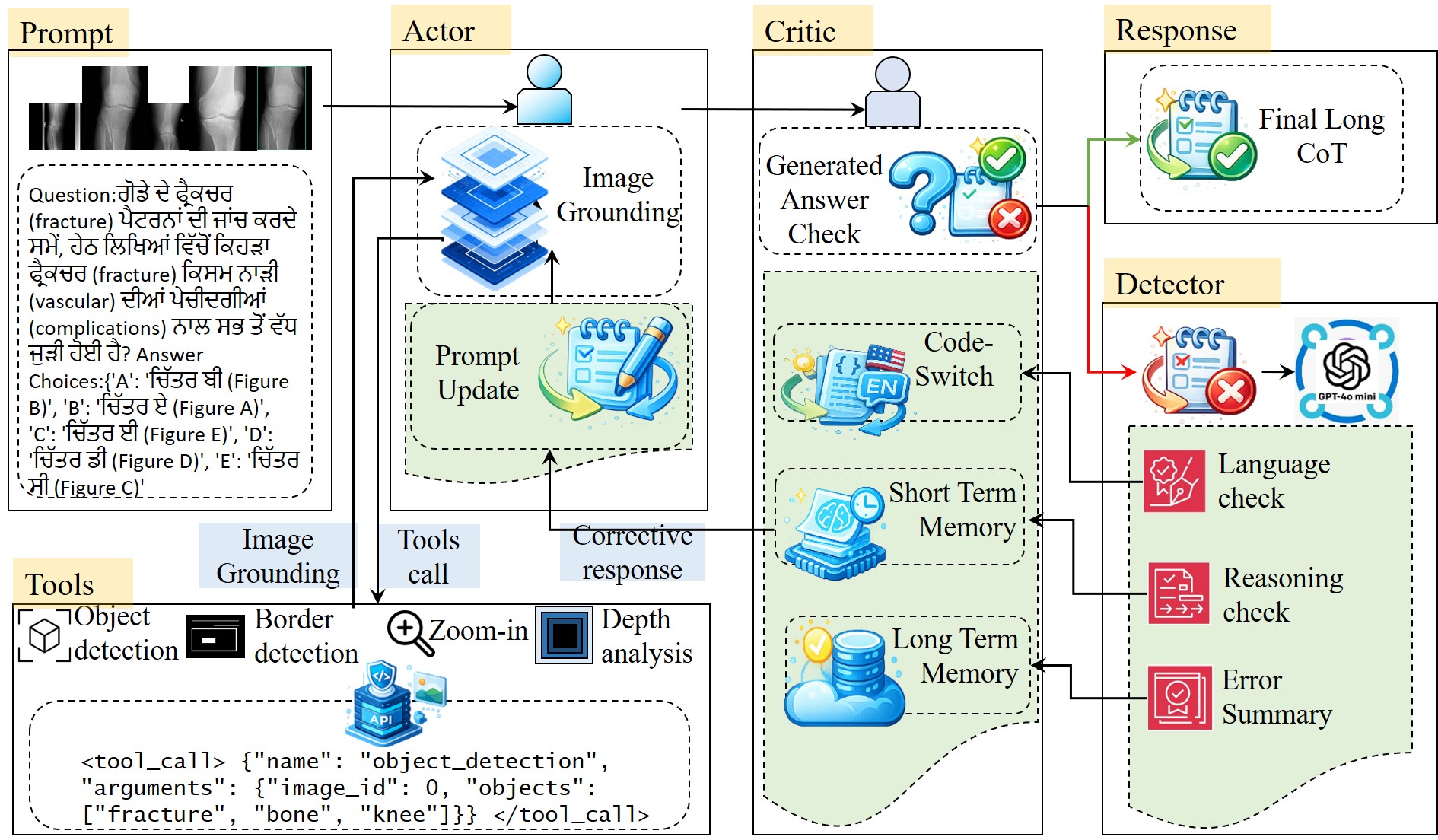}
    \caption{Overview of the \textbf{\textit{ArogyaSutra}} framework. ArogyaSutra employs an actor–critic architecture enhanced with tool-based image grounding and adaptive code-switching. The Actor first processes the input prompt and identifies the need for visual grounding, invoking appropriate tool agents to extract clinically relevant information from medical images before generating an answer with its associated reasoning. This output is then passed to the Critic for evaluation. If the response is correct, the Critic approves and outputs the final answer and reasoning. Otherwise, it consults an error detector (GPT-4o-mini) to identify the source of failure. Language-related errors trigger code-switching by translating the query into English, while reasoning-related errors are handled by incorporating summaries of past and current mistakes from long-term and short-term memory. The refined query is then fed back to the Actor for iterative refinement.}
    \label{fig:actor-critic}
\end{figure*}

To summarize, the contributions of this work can be enumerated as follows: \noindent\textbf{ Problem Formulation.}
We formalize a new task of \emph{multimodal medical reasoning in Indic languages}, where a model must perform step-by-step clinical reasoning over medical images and Indic-language queries to produce medically correct and linguistically consistent answers.\textbf{ Benchmark.} We introduce \textbf{ArogyaBodha}, a large-scale multilingual multimodal medical reasoning dataset and benchmark covering seven major Indian languages and English. Curated from eight diverse medical sources, it spans 31 body systems and 21 clinical domains, with expert-verified instances and unambiguous ground-truth answers to enable reliable evaluation of reasoning accuracy. \textbf{ Framework.}
We propose {\em \textbf{ArogyaSutra}}, an actor--critic-based multi-agent framework that combines tool-based visual grounding with dual-memory mechanisms for step-wise multimodal medical inference. The framework explicitly tracks and corrects reasoning errors and linguistic inconsistencies through iterative refinement. \textbf{ Evaluation.}
Extensive experiments across languages, imaging modalities, and clinical domains demonstrate that ArogyaSutra consistently outperforms strong multimodal baselines in both reasoning accuracy and multilingual alignment, highlighting its effectiveness for reliable medical AI in Indic language settings.

\begin{table}[t]
    \centering
    \small
    \begin{tabular}{clcp{3.2cm}}
        \toprule
        \textbf{Sl.} & \textbf{Dataset} & \textbf{Records} & \textbf{Processing} \\
        \midrule
        1 & MedXpertQA     & 1,982 & Direct \\
        2 & MedTrinity-25M & 576  & Caption, few-shot \\
        3 & MedPix-2.0     & 464  & Report, few-shot \\
        4 & MAMA-MIA       & 64   & Caption, few-shot, patient details \\
        5 & BRATS24        & 393  & Caption, few-shot, patient details \\
        6 & PMC-VQA        & 1,073 & Direct, filtered \\
        7 & GMAI-MMBench   & 164  & Direct, filtered \\
        8 & NEET-PG, FMGE  & 391  & Extracted, filtered \\
        \midrule
          & \textbf{Total} & \textbf{5107} & \\
        \bottomrule
    \end{tabular}
    \caption{Summary of datasets of English language after filtering with processing strategies.}
    \label{tab:dataset_summary}
\end{table}

\textbf{Impact for AI for Social Good.}
\textbf{\textit{ArogyaSutra}} and \textbf{ArogyaBodha} support equitable healthcare access by enabling multilingual multimodal medical reasoning in low-resource Indic languages. The work aligns with AI for Healthcare, Public Health, and Marginalized Communities by promoting clinically grounded and inclusive medical AI systems.

\textbf{Stakeholders and Multidisciplinary Collaboration.}
This work is developed in collaboration with certified medical practitioners for clinical validation of the curated medical questions and answers. Translation quality across seven Indic languages is evaluated using reverse-translation analysis with cosine similarity and BLEU$_4$ metrics to ensure preservation of clinical meaning and linguistic fidelity in multilingual healthcare settings.

\section{Related Works }
\subsection{Medical Multimodal and Multilingual Reasoning}
Recent training paradigms such as prompt tuning~\cite{Mohanty2025-rz}, supervised fine-tuning (SFT)~\cite{Huang2025-gt}, reinforcement learning (RL)~\cite{Zhou2025-kz} ,agentic distillation ~\cite{ghosh2026carepilot} have significantly improved MLLMs on complex vision--language reasoning tasks~\cite{Chen2023-fb} and have been increasingly adapted to medical settings~\cite{Huang2025-xn,Sun2025-af}. However, robust multimodal medical reasoning remains challenging due to limitations in supervision strategies and dataset quality~\cite{Zhang2025-pl,Hu2025-ec}. Prior studies show that SFT with final-answer supervision often leads to overfitting and poor out-of-distribution generalization, as observed in Medvlm-r1~\cite{Pan2025-nc}, while structured reasoning approaches such as Medvlthinker~\cite{Huang2025-xn} remain constrained by modality-specific data. RL-based and multi-agent methods, including Med-R1~\cite{Lai2025-it} and MMedAgent-RL~\cite{Xia2025-cg}, aim to improve robustness through long-form Chain-of-Thought (CoT) distillation, while Chiron-o1~\cite{Sun2025-af} achieves effective reasoning via long-CoT SFT with Mentor--Intern Collaborative Search. Multilingual medical reasoning poses additional challenges, as reasoning performance is strongest in English due to data imbalance~\cite{Qiu2024-tu}, and enforcing reasoning in low-resource languages often degrades coherence and amplifies degenerative language patterns~\cite{Li2025-vn}.

\subsection{Agentic Frameworks in Healthcare}
Recent work has explored agentic frameworks to enhance medical reasoning in complex clinical settings. MedAgents~\cite{tang2024medagents} introduces collaborative multi-agent discussion to improve zero-shot medical reasoning, while  MMedAgent~\cite{Li2024-sn} extends this paradigm using reinforcement learning to optimize coordination between specialist agents and a general practitioner agent. Med-R1~\cite{Lai2025-it} demonstrates that distilling Chain-of-Thought (CoT) reasoning via reinforcement learning improves robustness when high-quality medical supervision is limited. Complementary efforts such as AgentClinic ~\cite{schmidgall2024agentclinic} and MedAgentBench ~\cite{jiang2025medagentbench} focus on realistic evaluation environments for medical agents, emphasizing multi-step decision-making and tool interaction.~\cite{ghosh2026carepilot} improve long-horizon medical reasoning by introducing a dedicated dataset and a distillation-based framework for healthcare task automation. Despite these advances, most existing approaches are limited to high-resource languages or modality-specific settings. Our work builds upon these foundations by introducing an agentic, memory-aware multimodal reasoning framework tailored to low-resource Indic language healthcare scenarios. \textit{To address these gaps—particularly the lack of multimodal reasoning benchmarks for low-resource Indic languages—we curate a dedicated benchmark, \textbf{ArogyaBodha}. In addition, to enable agentic systems to make informed corrective decisions based on the underlying cause of failure, whether linguistic or logical, we propose a novel Actor–Critic framework, \textbf{\textit{ArogyaSutra}}, that adaptively governs feedback and reflection during reasoning.}

\section{Dataset: {\bf ArogyaBodha}}
\textbf{ArogyaBodha} comprises 7 major Indian languages namely
Bengali(5,111), 
Hindi(5,104), 
Assamese(5,108), 
Tamil(5,106), 
Telugu(5,106), 
Punjabi(5,108), 
and Marathi(5,107), 
 along with English(5,107)
curated and filtered from eight heterogeneous medical sources, resulting in a total of 40,857 samples. These sources include benchmark datasets, medical imaging repositories, and postgraduate medical examination questions from India.
\subsection{Curation and Filtration}
We build the dataset by combining existing medical reasoning benchmarks with questions from Indian postgraduate medical entrance examinations (NEET-PG and FMGE) as shown in Table (Table~\ref{tab:dataset_summary}). This ensures that the data is clinically relevant and covers a wide range of medical topics. To better reflect real-world clinical scenarios, we employ few-shot prompting with GPT-4o-mini by combining patient details, reports, and contextual medical information available in the datasets. To keep the questions challenging, The generated questions are further filtered based on reasoning depth and image–question relevance using Gemini-2.5-Flash. Relevance is evaluated using multiple clinical consistency criteria, including modality appropriateness (e.g., CT, MRI, X-ray), anatomical alignment between image and question, visibility of clinical findings, and overall clinical coherence between the image and the associated question. Only samples that pass the relevance checks are included in the final dataset. Finally, the clinical validity and medical correctness of the multimodal questions and answers are reviewed and verified by medical practitioners. 
\subsection{Translation}
The filtered high-quality questions and options are translated into seven major Indian languages using Gemini-2.5-Pro, which is one of the best-known LLMs for generating high-quality translations. This step enables multilingual coverage while preserving the original clinical semantics and reasoning requirements.


\subsection{Translation Analysis}
Translation quality is evaluated using reverse translation analysis  across seven Indic languages, as summarized in Table \ref{tab:translation_quality_question}. Semantic consistency between the original English text and translated medical questions is measured using cosine similarity with SentenceTransformer models, while BLEU$4$ scores are computed between the original and back-translated English text using NLTK. The results show strong semantic preservation across languages, with Cosine${Back}$ scores consistently remaining around 0.93--0.94 and higher direct semantic alignment observed for Hindi and Marathi.

To further assess translation quality, 20\% of the translated test samples are manually reviewed by medical practitioners proficient in the corresponding regional languages. The evaluators score the translations on a five-point scale based on clinical meaning preservation, medical terminology correctness, semantic consistency, linguistic fluency, and contextual relevance, resulting in an average score of 4.27. This combined automatic and human evaluation framework helps ensure linguistic fidelity, preservation of clinical meaning, and multilingual consistency throughout the benchmark construction process.

\begin{table}[t]
\centering
\small
\setlength{\tabcolsep}{5pt}
\renewcommand{\arraystretch}{1.1}
\begin{tabular}{lccc}
\toprule
\textbf{Language} & \textbf{BLEU$_4$} & \textbf{Cosine$_{Back}$} & \textbf{Cosine$_{Direct}$} \\
\midrule
\textbf{Hindi}     & 0.57 & 0.94 & \textbf{0.85} \\
Marathi            & 0.50 & 0.93 & 0.81 \\
Tamil              & 0.51 & 0.93 & 0.65 \\
Telugu             & 0.50 & 0.93 & 0.64 \\
Bengali            & 0.51 & 0.93 & 0.62 \\
Punjabi            & 0.56 & 0.93 & 0.58 \\
Assamese           & 0.49 & 0.93 & 0.50 \\
\bottomrule
\end{tabular}
\caption{
Reverse (back) translation quality for multilingual medical questions.
}
\label{tab:translation_quality_question}
\end{table}

\section{Framework : \textit{\textbf{ ArogyaSutra}}}
While recent multimodal models demonstrate strong perceptual grounding in generic environments, their performance degrades substantially in real-world healthcare systems, particularly in low-resource Indic language settings. To address these challenges, \textbf{\textit{ArogyaSutra}} introduces a multimodal agentic framework that integrates tool-grounded perceptual reasoning with dual-memory mechanisms comprising short-term and long-term memory to track prior errors and contextual dependencies across reasoning chains. In addition, \textbf{\textit{ArogyaSutra}} employs language-aware reflective reasoning to analyze past mistakes and dynamically determine when code-switching is required, enabling robust interaction with complex clinical software in Indic languages. The overall framework is shown in Figure \ref{fig:actor-critic}

\subsection{Vision Tools Grounding}
To enhance the perceptual grounding of medical findings, we focus on four visual tools that are particularly effective for perception-centric tasks. Although a broader set of tools is available, empirical observations show that the model consistently underutilizes them during inference. We therefore integrate these four high-performing tools directly into the MLLM inference loop, enabling tool execution and result incorporation to form a coherent multimodal rollout.  Specifically, the toolkit includes: (i) open-vocabulary \textit{object detection}, which localizes interface elements given a screenshot and textual query; (ii) \textit{zoom/crop}, which magnifies selected regions for fine-grained inspection; (iii) \textit{Edge Detection}, which extracts the edges of the objects present in the given image; and (iv) \textit{Depth Analysis}, which estimates the relative distances, producing a depth map.
\begin{table*}[t]
    \centering
    \small
    \begin{tabular}{lcccccccc}
        \toprule
        \textbf{Model} & \textbf{Assamese} & \textbf{Bengali} & \textbf{Hindi} & \textbf{Marathi} & \textbf{Punjabi} & \textbf{Tamil} & \textbf{Telugu} & \textbf{Average} \\
        \midrule
        GPT 4.0 mini & 35.38  & 45.38 &  37.69 &  33.07 &  36.15 &  31.53 & 33.20 &  36.05  \\ 
        GPT 4.0  & 38.40 & 44.60 & 40.10 & 37.80 & 39.20 & 36.90 & 39.10 & 39.30 \\

        Qwen3-VL-8B-Instruct                & 41.30 & 40.50 & 41.50 & 41.50 & 41.60 & 38.70 & 39.90 & 40.71 \\
        Mistral-Small-3.2-24B-Instruct      & 40.90 & 42.90 & \underline{44.10} & \underline{43.20} & 40.90 & 41.40 & \underline{42.50} & 42.27 \\
        LLaVA-v1.6-34B                      & 32.09 & 31.45 & 34.44 & 33.66 & 31.70 & 31.31 & 35.42 & 32.87 \\

        BioMistral-7B                      & 25.64 & 25.78 & 28.38 & 25.83 & 24.27 & 27.59 & 26.95 & 26.35 \\
        Llama3-OpenBioLLM-8B                & 16.10 & 15.60 & 22.20 & 20.50 & 11.70 & 12.10 & 7.70  & 15.13 \\
        MedGemma-4B-it                     & 38.50 & 35.10 & 37.40 & 34.50 & 35.70 & 36.70 & 34.90 & 36.11 \\
        MedVLM-R1                          & 23.50 & 22.80 & 25.60 & 22.80 & 24.20 & 25.40 & 22.20 & 23.79 \\
         \midrule
         Qwen2.5-VL-3B-Instruct              & 28.70 & 30.70 & 31.50 & 31.30 & 28.40 & 28.20 & 28.10 & 29.56 \\
        
 Qwen2.5-VL-7B-Instruct            & 32.30 & 36.50 & 33.84 & 36.12 & 32.30 & 35.38 & 33.07 & 34.21   \\
 \midrule
 \rowcolor{lightblue}
 \textbf{\textit{ArogyaSutra(Qwen 2.5 VL 3B)}}              & 34.69 & 37.08 & 38.36 & 37.03 & 33.07 & 35.73 & 33.60 & 35.65  \\
 \rowcolor{lightblue}
 \textbf{\textit{ArogyaSutra(Qwen 2.5 VL 7B)}}              & \underline{47.69} & \underline{45.38} & 42.30 & 42.30 & \underline{43.07} & \underline{41.53} & 41.53 & \underline{43.40}  \\
 
        \bottomrule
    \end{tabular}
    \caption{Comparison of \textbf{\textit{ArogyaSutra}} with respect to baselines on various Indic languages.
    The best results per column are highlighted in underline. The last group reports
the overall Average. * denotes our proposed method.}
    \label{tab:multilingual_model_comparison}
\end{table*}

\begin{table*}[t]
    \centering
    \small
    \begin{tabular}{lcccccccc}
        \toprule
        \textbf{Model} & \textbf{Assamese} & \textbf{Bengali} & \textbf{Hindi} & \textbf{Marathi} & \textbf{Punjabi} & \textbf{Tamil} & \textbf{Telugu} & \textbf{Average} \\
        \midrule
        \rowcolor{lightblue}
{\em \textbf{ArogyaSutra}(with all components)}              & 47.69 & 45.38 & 42.30 & 42.30 & 43.07 & 41.53 & 41.53 & 43.40  \\

w/o Critic and Image Grounding             & 40.00 & 36.00 & 34.00 & 32.00 & 32.00 & 28.00 & 32.00  & 33.43  \\
w/o Critic             & 42.00 & 32.0 & 36.00 & 30.00 & 34.00 & 30.00 & 32.00 & 33.71  \\
w/o Image Grounding              & 24.00 & 28.00 & 32.00 & 28.00 & 26.00 & 22.00 & 28.00 & 26.86  \\

w/o code-switch            & 30.00 & 32.00 & 26.00 & 24.00 & 28.00 & 24.00 & 24.00 & 26.86  \\
        \bottomrule
    \end{tabular}
    \caption{Results on various ablations in \textbf{\textit{ArogyaSutra}} show the impact of Image Grounding, critic agent, and code-switching.}
    \label{tab:aragyasutra_comparison}
\end{table*}

\subsection{Memory Utilization}
Reasoning errors may arise from linguistic ambiguities or incorrect intermediate logic errors. To prevent the agent from repeating such failures, we introduce an explicit memory mechanism that records past errors and guides future reasoning. The memory is structured into two components: a short-term memory that captures the most recent error, and a long-term memory that summarizes error patterns accumulated over the entire reasoning trajectory. Together, these components enable the agent to account for both immediate mistakes and persistent failure modes when making subsequent decisions. The short-term and long-term memories are represented by the equations below.

\begin{equation}
\mathcal{M}_{\text{short}}^{(t)} = \mathcal{E}_{t-1},
\end{equation}
\begin{equation}
\mathcal{M}_{\text{long}}^{(t)} =
\text{Summarize}\left(\{\mathcal{E}_0, \ldots, \mathcal{E}_{t-1}\}\right).
\end{equation}

\subsection{Actor-Critic Framework}
\textbf{\textit{ArogyaSutra}} adopts an Actor-Critic formulation as its core decision module, jointly leveraging tool-based perceptual grounding and temporally accumulated memory. Both the Actor and the Critic are instantiated from the same multimodal backbone, Qwen-VL-2.5-7B, and differ only in their functional roles : \textit{action proposal} versus \textit{action evaluation} and their input conditioning. At timestep $t$, the \textbf{Actor} observes the current interface state $o_t$, task objective $g$, perceptual context $z_t$, and memory states $(\mathcal{M}^{S}_t, \mathcal{M}^{L}_t)$, and samples a candidate semantic action:
\begin{equation}
u_t \sim \pi_{\theta}\!\left(u_t \mid o_t, g, z_t, \mathcal{M}^{S}_t, \mathcal{M}^{L}_t \right),
\end{equation}
where $\pi_{\theta}$ denotes the Actor policy and $u_t$ represents the proposed action. The \textit{Critic}, parameterized by $\psi$, evaluates the proposed action by estimating a scalar confidence score:
\begin{equation}
V_{\psi}\!\left(o_t, g, u_t, \mathcal{M}^{S}_t, \mathcal{M}^{L}_t \right) \rightarrow \hat{s}_t \in [0,1],
\end{equation}
where $\hat{s}_t$ reflects the estimated correctness of the action. If $\hat{s}_t$  = 1, the action is accepted, and both memory modules are updated; otherwise, the Critic triggers \textit{language-Analysis reflection} and emits structured corrective feedback $\Delta_t$ to guide the subsequent reasoning chain.

\textbf{Language Analysis Aware Reflection.}
When the Actor produces an incorrect prediction, the Critic first diagnoses the source of the failure, distinguishing between language understanding errors, such as repeated tokens or incomplete generations, and logical errors in the reasoning chain. If the error is attributed to language instability, the Critic provides reflective feedback in English to stabilize and guide the Actor’s reasoning chain to rectify itself. Conversely, if the failure stems from faulty reasoning, the Critic issues the reflection in the corresponding Indic language of the query, enabling more contextually grounded corrective guidance.

\textbf{Halting Condition.} If the actor–critic framework converges to the correct answer within three or fewer reasoning iterations, the resulting reasoning trace is retained. Otherwise, the reasoning process is reset and restarted using a summary of the accumulated errors stored in long-term memory. If the framework fails to reach the correct answer in two consecutive restart cycles, the corresponding data point is deemed unreliable and is removed from the training set.

\subsection{Training Strategy}
To progressively enhance the Actor’s capabilities, we perform code-switched reasoning chain distillation by transferring the Critic’s corrective signals into the Actor, thereby removing the need for explicit evaluation at inference time. Specifically, after simulating Actor-Critic rollouts, we construct a Critic-refined dataset 
$\mathcal{D} = \{(o_i, g_i, z_i, \mathcal{M}^{S}_i, \mathcal{M}^{L}_i, u_i^{\dagger})\}_{i=1}^{N}$,
where $u_i^{\dagger}$ denotes the Critic-approved action augmented with relevant tool outputs and memory context for the subsequent step.

The Actor is then fine-tuned using supervised learning with the following objective:
\begin{equation}
\mathcal{L}_{\text{SFT}}
= - \frac{1}{N} \sum_{i=1}^{N}
\log \pi_{\theta}\!\left(u_i^{\dagger} \mid o_i, g_i, z_i, \mathcal{M}^{S}_i, \mathcal{M}^{L}_i \right).
\end{equation}

During inference, only the distilled policy $\pi_{\theta}$ is retained. Conditioned on the current interface state, task objective, and memory signals, the Actor directly predicts the next semantic action without invoking the Critic. This distillation strategy preserves the Critic’s structured reasoning and memory-aware behavior while substantially reducing inference-time overhead.

\section{Experimental Setup}
This section describes our experimental setups. First section describes implementation details, second section describes evaluation metrics and the last section is for baselines.
\subsection{Implementation Details}
All experiments were conducted on an NVIDIA A100 80GB PCIe GPU. Models were trained for 3 epochs on an 18k multilingual multimodal dataset with balanced language proportions, requiring approximately 12--15 hours per run. We used PyTorch, Hugging Face Transformers, and Unsloth for efficient training. The learning rate was set to $5 \times 10^{-4}$ with 100 warmup steps and a weight decay of 0.01. Training was performed with a per-device batch size of 2 and 4 gradient accumulation steps. We fine-tuned the Qwen2.5-VL-7B-Instruct model using Unsloth with gradient checkpointing and 4-bit precision (\texttt{load\_in\_4bit=True}) for memory efficiency. Parameter-efficient fine-tuning was applied to the language, attention, and MLP projection layers, with LoRA rank $r=16$, LoRA $\alpha=16$, and dropout set to 0.

\subsection{Testset Composition}
From the curated dataset, we select 130 samples per language to construct the test set, resulting in 910 samples across seven languages. The test set is balanced across data sources, with identical questions used for all languages to enable controlled cross-lingual comparison. Each language’s test set comprises 42 samples from MedXpertQA, 34 from PMC-VQA, 16 from MedPix-2.0, 17 from MedTrinity-25M, 9 from BRATS24, 8 from NEET-PG and FMGE, 3 from GMAI-MMBench, and 1 from MAMA-MIA. We curate a set of 60 multimodal medical queries derived from real-world clinical examinations of the Médico Interno Residente (MIR) residency exams in Spain~\cite{OOD}, each translated into all seven evaluated Indic languages to enable controlled cross-lingual out-of-distribution (OOD) evaluation.

\begin{table}
    \centering
    \small
    \begin{tabular}{lc}
        \toprule
        \textbf{Model}  & \textbf{Accuracy} \\
        \midrule
        Qwen3-VL-8B-Instruct                &  49.6 \\
        Qwen2.5-VL-3B-Instruct              &  38.6 \\
        LLaVA-v1.6-34B                      &  23.3\\
       
        BioMistral-7B                      &  36.9 \\
        Llama3-OpenBioLLM-8B                &  39.3 \\
        MedGemma-4B-it                     &  45.2 \\
        MedVLM-R1                          &  23.6 \\
 \midrule
 Qwen2.5-VL-7B-Instruct (Baseline)              &  35.0  \\
\rowcolor{lightblue}
\textbf{ArogyaSutra(Qwen 2.5 VL 7B)}        & \textbf{50.4}  \\
        \bottomrule
    \end{tabular}
    \caption{Performance of \textbf{\textit{ArogyaSutra}} on OOD dataset.}
    \label{tab:ODD}
\end{table}

\subsection{Baselines and Evaluation Metric}
We evaluate the proposed question set across a diverse range of medical and non-medical multimodal large language models (MLLMs) to ensure a comprehensive comparison. Our evaluation includes vision-language models from the Qwen family~\cite{wang2024qwen2} (Qwen3-VL-8B-Instruct and Qwen2.5-VL-3B-Instruct) to analyze performance across model scales, as well as strong general-purpose MLLMs- Mistral-Small-3.2-24B-Instruct~\cite{jiang2024mixtral}, and LLaVA-v1.6-34B~\cite{liu2023visual} which are not specifically trained for medical reasoning. For domain-focused assessment, we additionally, benchmark medical MLLMs, including BioMistral-7B~\cite{labrak2024biomistral}, Llama3-OpenBioLLM-8B~\cite{dorfner2024biomedical}, MedGemma-4B-it~\cite{sellergren2025medgemma}, and MedVLM-R1~\cite{Pan2025-nc}. Among closed source models, we use GPT 4.0 mini and GPT 4.0. All models are evaluated in a zero-shot setting, and performance is measured using choice accuracy as the primary metric.

\begin{figure*}[t]
    \centering
    \includegraphics[width=0.95\textwidth]{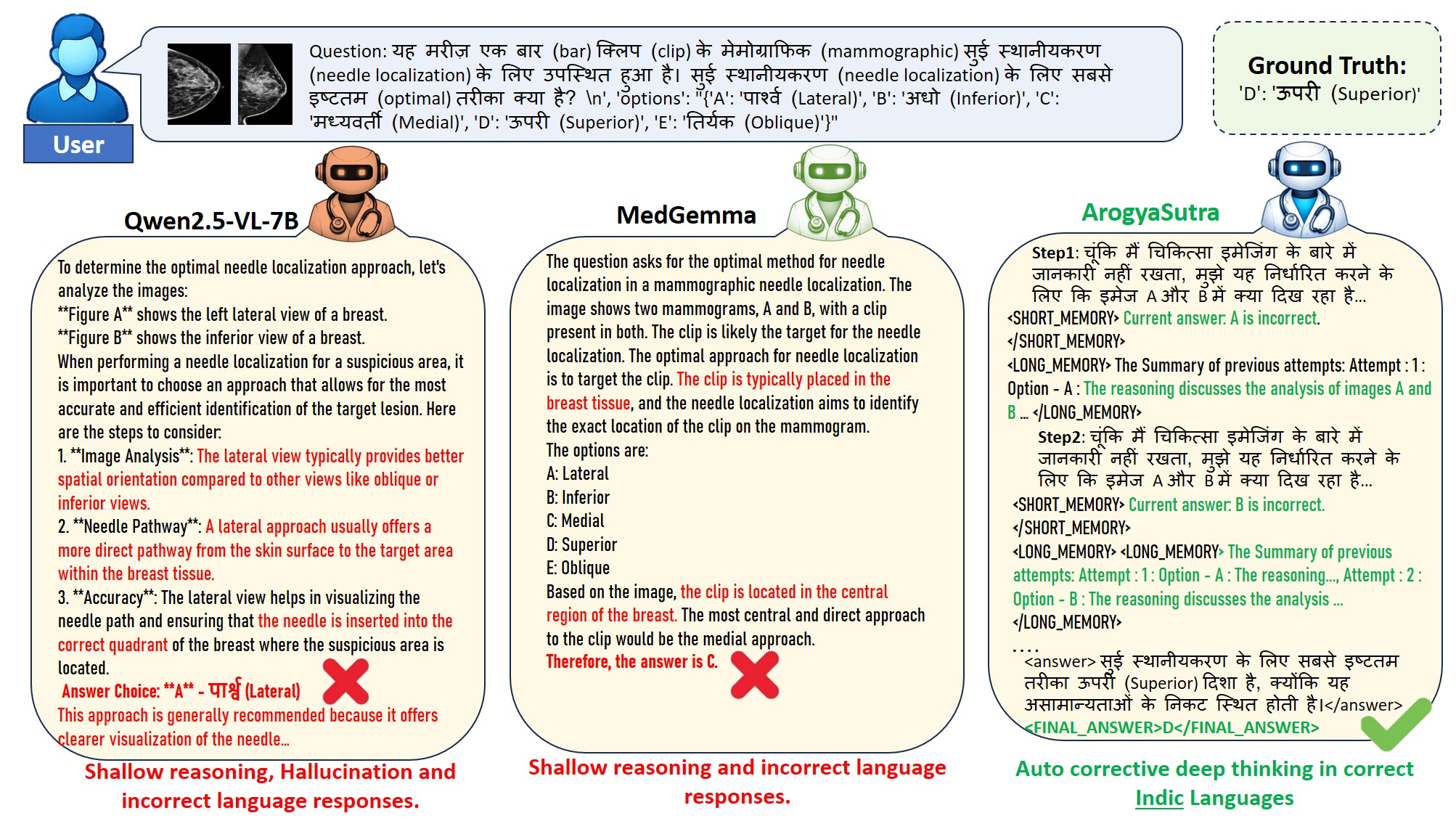}
    \caption{Overview of qualitative comparisons between \textbf{ \textit{ArogyaSutra}} and baseline models.}
    \label{fig:qualiative}
\end{figure*}

\section{Result and Analysis}
We evaluate \textbf{\textit{ArogyaSutra}} against strong \emph{general-purpose} and \emph{medical-domain} multimodal baselines. As shown in Table~\ref{tab:multilingual_model_comparison}, integrating tool grounding with short- and long-term memory substantially improves multimodal medical reasoning. We further conduct controlled component and robustness analyses on an out-of-distribution  benchmark ~\cite{OOD} medical questions across multiple Indic languages.

\subsection{Research Question}
\textbf{a) How does \textit{ArogyaSutra} perform compared to baselines?} Table  \ref{tab:multilingual_model_comparison}  shows that \textit{\textbf{ArogyaSutra}} achieves the best average performance across evaluated languages. The Qwen2.5-VL-7B variant achieves the highest average accuracy of \textbf{43.40}, surpassing the strongest baseline (GPT-4.0 at 39.30) by \textbf{+4.1 points}, with consistent gains across all seven languages. It also significantly outperforms medical-specific models such as \textit{BioMistral-7B} (23.83 avg) and \textit{MedGemma-4B-it} (36.11 avg), highlighting a clear advantage over domain-specialized baselines. Notably, \textbf{\textit{ArogyaSutra}} improves over its base Qwen2.5-VL-7B model by \textbf{+9.2 points}, demonstrating the effectiveness of the proposed approach for robust cross-lingual medical reasoning.

\par \textbf{b) How is the performance of \textit{ArogyaSutra} compared to the baseline models it has been trained on?} In Table  \ref{tab:multilingual_model_comparison},
\textit{\textbf{ArogyaSutra}} substantially outperforms the baseline models on which it is trained. \textbf{\textit{ArogyaSutra}} (Qwen2.5-VL-3B) improves the average accuracy from 29.56 to 35.65, yielding a gain of \textbf{+6.1 points} over Qwen2.5-VL-3B-Instruct. More prominently, ArogyaSutra (Qwen2.5-VL-7B) achieves an average accuracy of \textbf{43.40}, surpassing its base Qwen2.5-VL-7B-Instruct model (34.21) by \textbf{+9.2 points}, with consistent improvements across all evaluated languages. This indicates that \textbf{\textit{ArogyaSutra}} delivers substantial gains beyond its underlying baseline models.

\par \textbf{c) How is the performance of \textbf{\textit{ArogyaSutra}} on out of distribution dataset?} Table \ref{tab:ODD} shows the performance of \textbf{\textit{ArogyaSutra}} on OOD dataset. On the out-of-distribution (OOD) dataset~\cite{OOD}, \textbf{\textit{ArogyaSutra}} demonstrates strong generalization performance. \textbf{\textit{ArogyaSutra}}(Qwen2.5-VL-7B) achieves an accuracy of \textbf{50.4}, outperforming its base model Qwen2.5-VL-7B-Instruct baseline (35.0) by a clear margin. It also surpasses large general-purpose MLLMs such as LLaVA-v1.6-34B (23.3), as well as medical-specific models including BioMistral-7B (36.9) and MedGemma-4B-it (45.2). These results indicate that \textbf{\textit{ArogyaSutra}} exhibits robust reasoning and improved resilience under distribution shift.

\section{ Ablation Studies}


\textbf{Impact of the Critic Agent and its Components.} Removing the Critic significantly reduces performance, with Actor-only variants achieving 33.43\%--33.71\% compared to \textbf{43.40\%} for the full \textit{ArogyaSutra} framework (Table~\ref{tab:aragyasutra_comparison}). Even the grounded Actor-only model underperforms the baseline \textit{Qwen2.5-VL-7B-Instruct} (34.21\%). Furthermore, removing tool grounding or code-switching causes the largest performance drop (16.54\%), while disabling long- and short-term memory reduces accuracy by 12.83\%. These results highlight the importance of reflective error correction, grounding, memory, and adaptive multilingual reasoning in \textit{ArogyaSutra}.

\textbf{Impact of Tool Grounding and Memory.} Table \ref{tab:groun_memory_ablation} shows that disabling both components reduces accuracy to 26.86\%. Memory alone improves performance to 30.57\%, while enabling both achieves the best accuracy of \textbf{43.40\%}, demonstrating their complementary role in multilingual multimodal reasoning.



\begin{table}[t]
\centering
\begin{tabular}{c c c}
\hline
\textbf{Image}   & \textbf{Memory}   & \textbf{Average}   \\
\textbf{Grounding} &   \textbf{Utilization} &   \textbf{Output (\%)} \\
\hline
\includegraphics[height=10pt]{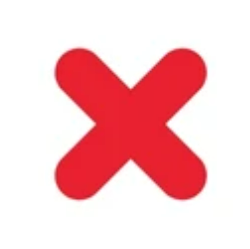} &
\includegraphics[height=10pt]{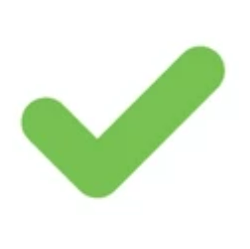} & 26.86 \\

\includegraphics[height=10pt]{img/check.png} &
\includegraphics[height=10pt]{img/cross.png} & 30.57 \\

\includegraphics[height=10pt]{img/check.png} &
\includegraphics[height=10pt]{img/check.png} & \textbf{43.40} \\
\hline
\end{tabular}
\caption{Ablation study of image grounding and memory utilization within the Critic agent. Best performance is achieved when both components are enabled.}
\label{tab:groun_memory_ablation}
\end{table}


\section{Qualitative Analysis}
We qualitatively analyze the generated outputs of the proposed model and compare them with two representative baselines: the medical MLLM \textit{MedGemma-4B-it} and the general-purpose vision--language model \textit{Qwen2.5-VL-7B-Instruct} as shown in the Figure \ref{fig:qualiative}. Both baselines exhibit shallow reasoning and frequently fail to respond in the input low-resource language (e.g., Hindi). In contrast, \textbf{\textit{ArogyaSutra}} consistently generates responses in the target language and demonstrates more structured and corrective reasoning. Leveraging the actor-critic framework, the model identifies and revises incorrect intermediate conclusions, progressively eliminating implausible hypotheses. These observations highlight the effectiveness of \textbf{\textit{ArogyaSutra}} in improving multilingual alignment and reasoning fidelity in multilingual medical settings.
\section{Risk Analysis and Limitations}
Despite the strong performance of \textbf{\textit{ArogyaSutra}} and the \textbf{ArogyaBodha} benchmark, several limitations remain. The model may still produce reasoning errors or misinterpret visual evidence, particularly in rare or atypical clinical cases, which poses risks if outputs are misused for real-world decision making. While \textbf{ArogyaBodha} covers seven major Indian languages, it does not capture the full linguistic diversity of India, including low-resource dialects and code-mixed language commonly used in clinical practice, potentially degrading performance under such conditions. Finally, the actor–critic framework depends on the reliability of underlying visual grounding tools and the base multimodal backbone; failures in perception or error attribution can propagate through the reasoning process despite critical feedback.

\section{Conclusion}

We introduce \textbf{ArogyaBodha} and \textbf{\textit{ArogyaSutra}} to advance multilingual multimodal medical reasoning in low-resource Indic-language settings, demonstrating consistent improvements over strong general-purpose and medical baselines across languages and imaging modalities. 
By integrating tool-grounded perception, actor--critic reasoning, memory-aware reflection, and adaptive code-switching, \textbf{\textit{ArogyaSutra}} enables more reliable and linguistically consistent medical reasoning across diverse Indic languages. Extensive experiments, ablation studies, and out-of-distribution evaluations demonstrate the effectiveness and robustness of the proposed framework. We hope this work encourages future research in trustworthy multilingual medical AI and clinically grounded multimodal reasoning systems for underserved communities.

\section*{Ethical Statement}
 All datasets are anonymized and contain no personally identifiable patient information. All datasets used in \textbf{ArogyaBodha}, including MedXpertQA, MedTrinity-25M, PMC-VQA, MedPix-2.0, MAMA-MIA, BraTS 2024, GMAI-MMBench, and NEET-PG/FMGE, are publicly available under their respective research, academic, or open-access licenses and are utilized solely for non-commercial research purposes.


\section*{Acknowledgments} The authors gratefully acknowledge the Aryabhatta Supercomputing Centre (ASC) at the Indian Institute of Technology Patna, established under the National Supercomputing Mission (NSM), Government of India, for providing the computational resources utilized in this work. The authors also sincerely thank Dr. Muhsin Muhsin from IGIMS and Dr. Priti Singh from IIT Patna for their valuable collaboration and support throughout the project, particularly in the data curation and verification processes.
\section*{Contribution Statement}
Tanmoy Kanti Halder and Akash Ghosh contributed equally to this work. 
\bibliographystyle{named}

\bibliography{ijcai26}

\end{document}